\documentclass{article}

\usepackage[final,nonatbib]{bdl_2018}

\usepackage[utf8]{inputenc} 
\usepackage[T1]{fontenc}    
\usepackage{hyperref}       
\usepackage{url}            
\usepackage{booktabs}       
\usepackage{amsfonts}       
\usepackage{nicefrac}       
\usepackage{microtype}      
\usepackage{graphicx}
\usepackage{blindtext}
\usepackage{multirow}
\usepackage{amsmath}
\newcommand{\JM}[1]{\textcolor{red}{JM:#1}}
\usepackage{algorithm}
\usepackage{algpseudocode}
\usepackage{todonotes}
\usepackage{changebar}
\title{Deep Probabilistic Ensembles: Approximate Variational Inference through KL Regularization}

\author{
	Kashyap Chitta\\
	Carnegie Mellon University\\
	\texttt{kchitta@andrew.cmu.edu} \\
	\And
	Jose M. Alvarez\\
	NVIDIA\\
	\texttt{josea@nvidia.com} \\
	\And
	Adam Lesnikowski \\
	NVIDIA \\
	\texttt{alesnikowski@nvidia.com} \\
}

\begin{document}
	
	\maketitle
	
	\begin{abstract}
		In this paper, we introduce Deep Probabilistic Ensembles (DPEs), a scalable technique that uses a regularized ensemble to approximate a deep Bayesian Neural Network (BNN). We do so by incorporating a KL divergence penalty term into the training objective of an ensemble, derived from the evidence lower bound used in variational inference. We evaluate the uncertainty estimates obtained from our models for active learning on visual classification. Our approach steadily improves upon active learning baselines as the annotation budget is increased.
	\end{abstract}
	
	\section{Introduction}
	Modeling uncertainty for deep neural networks has a wide range of potential applications: it can provide important information about the reliability of predictions, or better strategies for labeling data in order to improve performance. Bayesian methods, which provide an approach to do this, have recently gained momentum, and are beginning to find more widespread use in practice \cite{graves2011practical,blundell2015weight,gal2015bayesian,kendall2017uncertainties}.
	
	The formulation of a Bayesian Neural Network (BNN) involves placing a prior distribution over all the parameters of a network, and obtaining the posterior given the observed data \cite{neal1995bayesian}. The {distribution of predictions} provided by a trained BNN helps capture the model's uncertainty. However, training a BNN involves marginalization over all possible assignments of weights, which is intractable for deep BNNs without approximations \cite{graves2011practical,blundell2015weight,gal2015bayesian}. Existing approximation algorithms limit their applicability, since they do not specifically address the fact that deep BNNs on large datasets are more difficult to optimize than deterministic networks, and require extensive parameter tuning to provide good performance and uncertainty estimates \cite{osband2016risk}. Furthermore, estimating uncertainty in BNNs requires drawing a large number of samples at test time, which can be extremely computationally demanding.
	
	In practice, a common approach to estimate uncertainty is based on ensembles \cite{lakshminarayanan2017simple,beluch2018power}. Different models in a trained ensemble of networks are treated as if they were samples drawn directly from a BNN posterior. Ensembles are easy to optimize and fast to execute. However, they do not approximate uncertainty in the same manner as a BNN. For example, the parameters in the first kernel of the first layer of a convolutional neural network may serve a completely different purpose in different members of an ensemble. Therefore, the variance of the values of these parameters after training cannot be compared to the variance that would have been obtained in the first kernel of the first layer of a trained BNN with the same architecture.
	
	In this paper, we propose Deep Probabilistic Ensembles (DPEs), a novel approach to approximate BNNs using ensembles. Specifically, we use variational inference \cite{blei2016variational}, a popular technique for training BNNs, to derive a KL divergence regularization term for ensembles. DPEs are parallelizable, easy to implement, yield high performance, and in our experiments, provide better uncertainty estimates than existing methods when used for active learning on visual classification benchmarks.
	
	\section{Deep Probabilistic Ensembles}
	
	For inference in a BNN, we can consider the weights $w$ to be latent variables with some prior distribution, $p(w)$. These weights relate to the observed dataset $x$ through the likelihood, $p(x|w)$. We aim to compute the posterior $p(w|x)$ that best explains the observed data. Variational inference involves restricting ourselves to a family of distributions $D$ over the latent variables, and optimizing for the member of this family $q^*(w)$ that is closest to the true posterior in terms of KL divergence, 
	\begin{align}
		q^*(w) & = \underset{q(w) \in D}{\arg\min} \ KL(q(w)||p(w|x)).
	\end{align}
	Simplifying this objective, we obtain the negative Evidence Lower Bound (ELBO) \cite{blei2016variational},
	\begin{align}
		-ELBO & = KL(q(w)||p(w)) - \mathbb{E}[\log p(x|w)],
		\label{e:elbo}
	\end{align}
	where the first term is the KL divergence between $q(w)$ and the chosen prior distribution $p(w)$; and the second term is the expected negative log likelihood (NLL) of the data $x$ based on the current parameters $w$. The optimization difficulty in variational inference arises partly due to this expectation of the NLL. In deterministic networks, fragile co-adaptations exist between different parameters, which can be crucial to their performance \cite{yosinski2014transferable}. Features typically interact with each other in a complex way, such that the optimal setting for certain parameters is highly dependent on specific configurations of the other parameters in the network. Co-adaptation makes training easier, but can reduce the ability of deterministic networks to generalize \cite{hinton2012improving}. Popular regularization techniques to improve generalization, such as Dropout, can be seen as a trade-off between co-adaptation and training difficulty \cite{srivastava2014dropout}. An ensemble of deterministic networks exploits co-adaptation, as each network in the ensemble is optimized independently, making them easy to optimize.
	
	For BNNs, the nature of the objective, an expectation over all possible assignments of $q(w)$, prevents the BNN from exploiting co-adaptations during training, since we seek to minimize the NLL for \textit{any generic deterministic network} sampled from the BNN. While in theory, this should lead to great generalization, it becomes very difficult to tune BNNs to produce competitive results. In this paper, we propose a form of regularization to use the optimization simplicity of ensembles for training BNNs.
	
	\noindent
	\textbf{KL regularization.} The standard approach to training neural networks involves regularizing each individual parameter with $L_1$ or $L_2$ penalty terms. We instead apply the KL divergence term in Eq. \ref{e:elbo} as a regularization penalty $\Omega$, to the \textit{set of values} of a given parameter takes over \textit{all members in an ensemble}. If we choose the family of Gaussian functions for $p(w)$ and $q(w)$, this term can be analytically computed by assuming mutual independence between the network parameters and factoring the term into individual Gaussians. The KL divergence between two Gaussians with means $\mu_q$ and $\mu_p$, standard deviations $\sigma_q$ and $\sigma_p$ is given by
	\begin{equation}
		KL(q||p) = \frac{1}{2}\left(\log \frac{\sigma_q^2}{\sigma_p^2} + \frac{\sigma_p^2 + (\mu_q - \mu_p)^2}{\sigma_q^2} - 1\right).
		\label{e:kl}
	\end{equation}
	
	\textbf{Choice of prior.} In our experiments, we choose the network initialization technique proposed by He et al. \cite{he2015delving} as a prior. For batch normalization parameters, we fix $\sigma_p^2=0.01$, and set $\mu_p=1$ for the weights and $\mu_p=0$ for the biases. For the weights in convolutional layers with the ReLU activation (with $n_i$ input channels, $n_o$ output channels, and kernel dimensions $w \times h$) we set $\mu_p=0$ and $\sigma_p^2=\frac{2}{n_{o}wh}$. Linear layers can be considered a special case of convolutions, where the kernel has the same dimensions as the input activation. The KL regularization of a layer $\Omega^l$ is obtained by removing the terms independent of $q$ and substituting for $\mu_p$ and $\sigma_p$ in Eq. \ref{e:kl},
	\begin{equation}
		\Omega^l = \sum_{i=1}^{n_{i} n_{o} w h} \left(\log {\sigma_i^2} + \frac{2}{n_{o} w h\sigma_i^2} + \frac{\mu_i^2}{\sigma_i^2}\right),
		\label{e:omega}
	\end{equation}
	\noindent where $\mu_i$ and $\sigma_i$ are the mean and standard deviation of the set of values taken by a parameter across different ensemble members. In Eq. \ref{e:omega}, the first term prevents extremely large variances compared to the prior, so the ensemble members do not diverge completely from each other. The second term heavily penalizes variances less than the prior, promoting diversity between members. The third term closely resembles weight decay, keeping the mean of the weights close that of the prior, especially when their variance is also low.
	
	\textbf{Objective function.} We can now rewrite the minimization objective from Eq. \ref{e:elbo} for an ensemble as:
	\begin{equation}
		\mathbf{\Theta}^* = \underset{\mathbf{\Theta}}{\arg\min} \sum_{i=1}^{N} \sum_{e=1}^{E} \mathcal{H}(y^i,\mathcal{M}_e(\textbf{x}^i,\Theta_e)) + \beta \Omega( \mathbf{\Theta}),
	\end{equation}
	where $\{(\textbf{x}^i,y^i)\}_{i=1}^{N}$ is the training data, $E$ is the number of models in the ensemble, and ${\Theta}_e$ refers to the parameters of the model $\mathcal{M}_e$. $\mathcal{H}$ is the cross-entropy loss for classification and $\Omega$ is our KL regularization penalty over all the parameters of the ensemble $\mathbf{\Theta}$. We obtain $\Omega$ by aggregating the penalty from Eq. \ref{e:omega} over all the layers in a network, and use a scaling term $\beta$ to balance the regularizer with the likelihood loss. By summing the loss over each independent model, we are approximating the expectation of the ELBO's NLL term in Eq. \ref{e:elbo} over only the current ensemble configuration, a subset of all possible assignments of $q(w)$. This is the main distinction between our approach and traditional variational inference.
	
	\section{Experiments}
	
	\begin{table}[t]
		\begin{center}
			\caption{Validation accuracies (in \%) on an active learning task. DPEs give consistent improvements in results over both random sampling and standard ensembles on both datasets. Relative performance to the upper bound performance for that data sampling strategy is given in (brackets).}
			\label{t:active_32}
			\begin{tabular}{c|c|c|c|c}
				\hline
				\textbf{Task} & \textbf{Data Sampling} & \textbf{Accuracy @8\%} & \textbf{Accuracy @16\%} & \textbf{Accuracy @32\%}\\
				\hline
				& Random & 80.60 (84.66) & 86.80 (91.18) & 91.08 (95.67) \\
				CIFAR-10 & Ensemble & 82.41 (86.56) & 90.05 (94.59) & 94.13 (98.87)\\
				& DPE (Ours) & \textbf{82.88 (87.06)} & \textbf{90.15 (94.70)} & \textbf{94.33 (99.09)} \\
				\hline
				& Random & 39.57 (50.18) & 54.92 (69.64) & 66.65 (84.51) \\
				CIFAR-100 & Ensemble & 40.49 (51.34) & 56.89 (72.14) & 69.68 (88.36) \\
				& DPE (Ours) & \textbf{40.87 (51.83)} & \textbf{56.94 (72.20)} & \textbf{70.12 (88.92)} \\
				\hline
			\end{tabular}
		\end{center}	
	\end{table}
	
	Active learning allows a model to choose the data from which it learns, by iteratively using the partially trained model to {decide which examples to annotate} from a large unlabeled dataset \cite{cohn1994active}. In our approach, the data with the highest {entropy} for the {average prediction} across the ensemble members is added to the training set from the unlabeled pool. We experiment with active learning on the CIFAR dataset, which has two object classification tasks over natural images: one coarse-grained over 10 classes and one fine-grained over 100 classes \cite{krizhevsky2009learning}. There are 50k training images and 10k validation images of resolution $32 \times 32$.
	
	We use eight models in our ensembles, each a pre-activation ResNet-18 \cite{he2016deep}. Our first experiment involves initializing models by training with a random subset of 4\% of the dataset, and then re-training the model at three additional intervals after adding more data (at 8\%, 16\% and 32\% of the data). We evaluate three approaches for adding data to the training set, (1) Random: pick the required percentage of data through random sampling; (2) Ensemble: pick the samples with highest average entropy across our eight models with standard $L_2$ regularization for each model; and (3) Deep Probabilistic Ensemble (DPE): pick samples with the highest average entropy across our eight models jointly regularized with our KL regularization. Our results are shown in Table \ref{t:active_32}. Our figures correspond to the mean accuracy of 3 experimental trials. As shown, active learning with DPEs clearly outperforms random baselines, and consistently provides better results compared to ensemble based active learning methods. Further, though the gap is small, it typically grows as the annotation budget (size of the labeled dataset) increases, which is a particularly appealing property for large-scale labeling projects.
	
	In our second experiment, we compare our approach to state-of-the-art results on the CIFAR-10 dataset using 20\% of the labeled training data. We include results for (1) the proposed DPE; (2) a deterministic network using the same architecture and $L_2$ regularization; (3) Core-set selection with a single VGG-16~\cite{sener2018active} and; (4) an ensemble of DenseNet-121 models~\cite{beluch2018power}. We also include the upper bound of each experiment. That is, the accuracy obtained when the model is trained using all the data available. For the first two methods, we start training at 4\% of the data and then re-train the model at four additional intervals (at 8\%, 12\%, 16\% and 20\%). For the others we use the performance reported in the corresponding papers. As shown in Table \ref{t:compare}, our approach outperforms all the others, not only achieving higher accuracy with limited training data, but also reducing the gap to the corresponding upper bound.
	
	\begin{table}[t]
		\begin{center}
			\caption{\textbf{CIFAR-10:} Comparison of our proposed approach to active learning baselines for learning with limited labels. DPEs show state-of-the-art performance. Relative performance refers to the performance for each method using a limited labeling budget compared to the performance using the whole dataset.}
			\label{t:compare}
			\begin{tabular}{c|c|c|c}
				\hline
				\textbf{Method} & \textbf{{Accuracy @20\%}} & \textbf{{Accuracy @100\%}} & \textbf{Relative Performance}\\
				\hline
				Core-set \cite{sener2018active} & 74\% & 90\% & 82.2\%\\
				Ensemble \cite{beluch2018power}  & 85\% & \textbf{95.5\%} &89.0\% \\
				\hline
				Deterministic Network& 87.5\% & 94.4\% & 92.7\%\\
				DPE (Ours) & \textbf{92\%} & 95.2\% & \textbf{96.3\%}\\
				\hline
			\end{tabular}
		\end{center}	
	\end{table}
	
	\section{Conclusion}
	
	In this paper, we introduced Deep Probabilistic Ensembles (DPEs) for uncertainty estimation in deep neural networks. The key idea is to train ensembles with a novel KL regularization term as a means to approximate variational inference for BNNs. Our results demonstrate that DPEs improve performance on active learning tasks over baselines and state-of-the-art active learning techniques on two image classification datasets. Importantly, contrary to traditional Bayesian methods, our approach is simple to integrate in existing frameworks and scales to large models and datasets without impacting performance. We look forward to future work with these models for any downstream tasks requiring precise uncertainty information.  
	
	\small
	\bibliographystyle{plain}
	\bibliography{references}
	
\end{document}